\documentclass[letterpaper, 10 pt, conference]{ieeeconf}  % Comment this line out if you need a4paper
\usepackage{times}
\usepackage{epsfig}
\usepackage{graphicx}
\usepackage{amsmath}
\usepackage{amssymb}
\usepackage{float}
\usepackage{relsize}
\usepackage{epstopdf}

\usepackage{booktabs}
\usepackage{pdfpages}
\usepackage{graphics}
\usepackage{subcaption}
\usepackage{siunitx}
\usepackage{hyperref}
\usepackage{booktabs}
\newcommand{\etal}{\mbox{\emph{et al.\ }}}

\IEEEoverridecommandlockouts                              % This command is only needed if 
\title{\LARGE \bf
\emph{Disparity Sliding Window}: Object Proposals From Disparity Images 
}

\author{Julian M\"uller$^{1}$, Andreas Fregin$^{2}$ and Klaus Dietmayer$^{1}$% <-this % stops a space
\thanks{*This work was supported by Daimler AG, Ulm}% <-this % stops a space
\thanks{$^{1}$Julian M\"uller and Klaus Dietmayer are with the Department of Measurement, Control and Microtechnology, University of Ulm, 89081 Ulm, Germany
        {\tt\small\{julian-2.mueller, klaus.dietmayer\}@uni-ulm.de}}%
\thanks{$^{2}$Andreas Fregin is with Daimler AG, Research and Development, 89081 Ulm, Germany
        {\tt\small andreas.fregin@daimler.com}}%
}

\begin{document}

\maketitle
\thispagestyle{empty}
\pagestyle{empty}

%%%%%%%%%%%%%%%%%%%%%%%%%%%%%%%%%%%%%%%%%%%%%%%%%%%%%%%%%%%%%%%%%%%%%%%%%%%%%%%%
\begin{abstract}
Sliding window approaches have been widely used for object recognition tasks in recent years~\cite{Papageorgiou1998,emlo,hog,pc}. They guarantee an investigation of the entire input image for the object to be detected and allow a localization of that object. Despite the current trend towards deep neural networks, sliding window methods are still used in combination with convolutional neural networks~\cite{Sermanet2013}. The risk of overlooking an object is clearly reduced compared to alternative detection approaches which detect objects based on shape, edges or color. Nevertheless, the sliding window technique strongly increases the computational effort as the classifier has to verify a large number of object candidates.\\
This paper proposes a sliding window approach which also uses depth information from a stereo camera. This leads to a greatly decreased number of object candidates without significantly reducing the detection accuracy. A theoretical investigation of the conventional sliding window approach is presented first. Other publications to date only mentioned rough estimations of the computational cost. A mathematical derivation clarifies the number of object candidates with respect to parameters such as image and object size. Subsequently, the proposed disparity sliding window approach is presented in detail.\\
The approach is evaluated on pedestrian detection with annotations and images from the KITTI~\cite{Geiger2012} object detection benchmark. Furthermore, a comparison with two state-of-the-art methods is made. Code is available in C++ and Python \url{https://github.com/julimueller/disparity-sliding-window}.    
\end{abstract}

%%%%%%%%% BODY TEXT
\section{Introduction}

Object recognition is a task in computer vision which occupies researchers in many different fields of application. Typically objects are detected by first detecting an amount object proposals. Those object proposals, or also called region proposals, are verified by a subsequent classification step. Early approaches are based on sliding-window technique. Objects of different size are detected with sub-window scalings at each position in the image. Despite modern computing power, sliding window approaches have limited real-time capability. The conventional implementation is computationally expensive as objects have to be searched with windows of different \emph{positions}, \emph{scales} and \emph{aspect ratios}.\\
The key question of how objects proposals can be defined and located was discussed by~\cite{Alexe2010} as one of the first. In the following years, a lot of methods for generating object proposals arised. Those approaches are often based on segmentation, edges, saliency or superpixels. However, object proposal generation still consumes more time than the classification step.\\
With rising success of convolutional neural networks for proposal classification, methods which predict object proposals from convolutional networks were published. Often, they share convolutional layers between the proposal and classification task, which makes proposal generation cheap. Nevertheless, high hardware and power requirement makes such approaches unattractive for many applications. Despite high parallelization applied on expensive and power-consuming GPUs, CNNs struggle to be real-time capable on higher image resolutions.\\
In this paper, we present a simple, yet efficient method for calculating object proposals from disparity images in a sliding-window fashion. We theoretically prove the dilemma of conventional sliding-window, which creates a huge amount of object proposals. Our disparity sliding window approach (DSW) is based on disparity images. Due to the additional depth information, one can predict the size of the bounding boxes at each position via pinhole camera model. Therefore, a 3D model of the object has to be defined. This avoids the generation of multiple sizes per image position compared to conventional sliding-window. Additionally, the window step size can be calculated adaptively. Close object regions lead to higher step sizes, whereas far objects have to be detected with small step sizes. We prove this in a theoretical manner. The created boxes are proved for their disparity homogenity to remove further inappropriate candidates.  Precalculated lookup-tables help to create the proposals in real-time. We evaulate our method on pedestrian detection using the well-established KITTI~\cite{Geiger2012} dataset.      

\section{Related work}

\textbf{Sliding Window Approches.} Acceleration methods were published in order to counter the computational cost of sliding window approaches. Lampert \etal 
introduced an efficient subwindow search~\cite{bsw} maximizing a large class of classifier functions over all subwindows. Vedaldi \etal presented a multiple kernel classifier for object detection~\cite{mkod} consisting of a cascade classifier approach with rising classifier complexity. However, their resulting system still requires around 60 seconds computation time due to the high number of sliding window objects. Among the first, Viola \etal~\cite{vj1,vj2} were able to detect faces in real-time. For this purpose fast features such as haar-features~\cite{Papageorgiou1998,Lienhart02anextended} were necessary to allow classifying thousands or millions of objects in a short time. Other approaches use heuristics and constraints~\cite{eugc} to reduce the number of candidates per frame. Using heuristics is critical as the risk of overlooking objects increases. Wojek \etal \cite{wojek08dagmb} speed up sliding window approaches using a GPU implementation. Song \etal~\cite{DBLP:conf/eccv/SongX14} slide a 3D window in depth maps and detect objects via SVM classifier. Detectors which detect objects based on shape~\cite{conf/cvpr/MajiM09} or color~\cite{Gevers97colorbased} are vulnerable to changes in viewpoint or illumination.\\
\textbf{Object Proposal Approches.} Alexe et al. \cite{Alexe2010,Alexe2012} rank a high number of windows per image by using an objectness score based on color, superpixels, edges and saliency. Selective Search~\cite{Uijlings2012}, one of the most poular approaches, returns a number of windows based on multiple hierarchical segmentations using superpixels from all color channels. Further segmentation-based approaches are~\cite{Carreira2010} and~\cite{Endres}. Another popular approach is EdgeBoxes~\cite{Zitnick2014}, a fast method which predicts object proposals from edges in less than a second. \\
\textbf{Object Proposals from Deep Learning.} OverFeat~\cite{Sermanet2013} detects bounding box coordinates from a fully-connected layer. A more general approach is MultiBox~\cite{Erhan2014} predicting bounding boxes for multi-class tasks from a fully-connected layer. YOLO~\cite{Redmon} predicts bounding boxes from a single layer, whereas classification and proposal generation share the same base network. SSD~\cite{Liu2015} enhances this approach by predicting bounding boxes from different layers. Faster R-CNN~\cite{Ren2015} simultaneously predicts object bounds and objectness scores at each position.

%------------------------------------------------------------------------
\section{The Dilemma of Conventional Sliding Window Approaches}
\label{sec:csw}
Because only rough estimations of the computational effort required for the sliding window method have been recently published~\cite{bsw,mkod}, the following analyzes the conventional sliding window approach. 

\subsection{Theoretical Analysis}
\begin{figure}[!t]
	\begin{center}
		\includegraphics[width=0.9\linewidth]{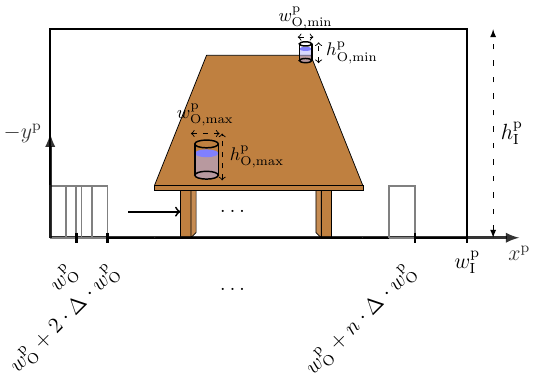}
	\end{center}
	\caption{Object detection problem with an image size of $w_\mathrm{I}^\mathrm{p}\times h_\mathrm{I}^\mathrm{p}$. Objects with a minimum/maximum width of $w_{\mathrm{O,min}}^\mathrm{p}$ and $w_{\mathrm{O,max}}^\mathrm{p}$ shall be detected. A sliding window is shifted with a step size $\Delta$. $N_\mathrm{x}=n+1$ windows fit into one row.}
	\label{fig:dsw_parameters}
\end{figure}

Since the number of candidates is crucial for the real-time capability this knowledge is of great importance for the design of a sliding window detector.\\ 
Figure~\ref{fig:dsw_parameters} illustrates the most important terms necessary to mathematically describe sliding window approaches. Two identical beverage cans placed at different distances are used as the objects to be detected. A camera image with an image height $h^\mathrm{p}_\mathrm{I}$ and width $w^\mathrm{p}_\mathrm{I}$ is given. The superscript $p$ denotes pixel coordinates. Objects with a minimum width of $w^\mathrm{p}_{\mathrm{O,min}}$ and a maximum width of $w^\mathrm{p}_{\mathrm{O,max}}$ shall be detected.

\begin{figure}[!t]
 	\begin{center}
 		\includegraphics[width=0.6\linewidth]{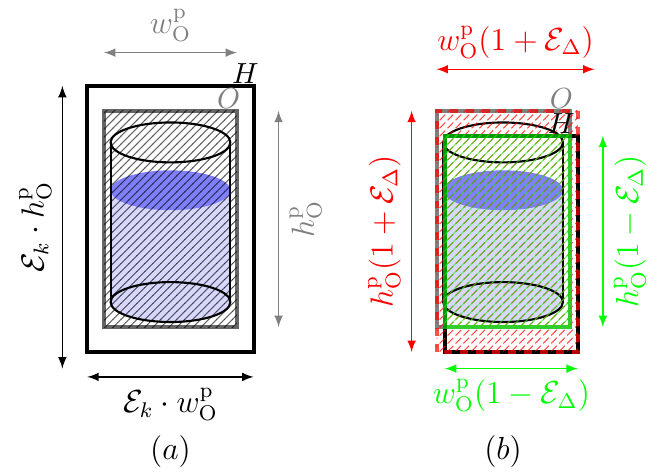}
 	\end{center}
 	\caption{The to be detected object is circumscribed by a bounding rectangle $O$. In order to detect the object accurate, an object candidate hypothesis $H$ has to meet two main conditions: scaling and positioning accuracy. Two examples with a scaling error (a) and position error (b) are illustrated.}
 	\label{fig:scaling_shifting}
 \end{figure}

The plot illustrates how a sliding window is typically shifted in $x^\mathrm{p}$ direction. The step size is chosen as a percentage $\Delta$ of the object candidate width $w^\mathrm{p}_{\mathrm{O}}$. Equation 
 
 \begin{align}
 w^\mathrm{p}_{\mathrm{O}} + n\cdot\Delta \cdot w^\mathrm{p}_{\mathrm{O}} \leq w^\mathrm{p}_{\mathrm{I}}
 \end{align}
 describes the condition, that the width of all horizontally aligned hypotheses must be smaller than or equal to the image width.
 Since the number of hypotheses $N_\mathrm{x}$ in $x^\mathrm{p}$ direction is given by $N_\mathrm{x} = n+1$, it can be calculated as
 \begin{align}
 N_\mathrm{x} = n + 1 \leq \frac{1}{\Delta}\Big(\frac{w^\mathrm{p}_\mathrm{I}}{w^\mathrm{p}_{\mathrm{O}}}-1\Big) +1 . 
 \end{align}
 The same assumption is used to calculate the number of hypotheses $N_\mathrm{y}$ in $y^\mathrm{p}$ direction assuming an aspect ratio $r$.
 The overall number of hypotheses $N$ for a fixed window width can then be formulated as
 \begin{align}
 N& \leq \Big( \frac{1}{\Delta}\Big(\frac{w^\mathrm{p}_\mathrm{I}}{w^\mathrm{p}_{\mathrm{O}}}-1\Big) +1 \Big)\cdot \Big( \frac{1}{\Delta}\Big(\frac{h^\mathrm{p}_\mathrm{I}}{r\cdot w^\mathrm{p}_{\mathrm{O}}}-1\Big) +1 \Big),
 \end{align} 
 where $N=N_\mathrm{x}\cdot N_\mathrm{y}$.
  When considering a variable window width $w^\mathrm{p}_{\mathrm{O,}i}$ and aspect ratio $r_i$, this formula extends to
 {\scriptsize
 \begin{align}
 N \leq \mathlarger{\sum}\limits_{w^\mathrm{p}_{\mathrm{O,min}}}^{w^\mathrm{p}_{\mathrm{O,max}}} ~\mathlarger{\sum}\limits_{r_{\mathrm{min}}}^{r_{\mathrm{max}}} \Big( \frac{1}{\Delta}\Big(\frac{w^\mathrm{p}_\mathrm{I}}{w^\mathrm{p}_{\mathrm{O,}i}}-1\Big) +1 \Big)  \Big( \frac{1}{\Delta}\Big(\frac{h^\mathrm{p}_\mathrm{I}}{r_i\cdot w^\mathrm{p}_{\mathrm{O,}i}}-1\Big) +1 \Big)
  \label{eq:hyps}
 \end{align}}
 In order to guarantee an accurate object detection, two conditions must be met. On the one hand, the window size $w^\mathrm{p}_{\mathrm{O,}i}$ has to be increased in appropriate steps from $w^\mathrm{p}_{\mathrm{O,min}}$ to $w^\mathrm{p}_{\mathrm{O,max}}$. On the other hand, the step size $\Delta$ must not be too large in order to not overlook objects. The overlap is crucial for a following classification step, because many classifiers are sensitive to position or scaling errors. How $w^\mathrm{p}_{\mathrm{O,}i}$ and $\Delta$ must be chosen to reach a defined accuracy is explained by deriving the scaling and positioning error. For easier computation we consider both conditions separately. 
 Figure~\ref{fig:scaling_shifting} illustrates an object to be detected circumscribed by an ideal bounding rectangle $O$ as well as inaccurate object hypotheses concerning their size by a scaling error $\mathcal{E}_k$ (a) and position by a positioning error $\mathcal{E}_\Delta$ (b). Such imperfect hypotheses are typically created by conventional sliding window approaches for a defined step size for the window.
 How the unknown parameters in Equation (\ref{eq:hyps}), step size $\Delta$ and the window scaling, beginning with $w^\mathrm{p}_{\mathrm{O,min}}$ up to $w^\mathrm{p}_{\mathrm{O,max}}$, have to be chosen must be determined. Both, step size of the window and scaling of the window size, are crucial to the detection accuracy. The relationship between the detection accuracy and those parameters is expressed using the metric intersection over union
 
 \begin{equation}
 \theta_{\mathrm{IOU}}= \frac{|H \cap O|}{|H \cup O|} = \frac{|H \cap O|}{|H| + |O| - |H \cap O|}.  
 \label{iou_theta}
 \end{equation} 
 In the following we approximate the union value by the area of a rectangle circumscribing both boxes (red area in Figure~\ref{fig:scaling_shifting}).
  Most object detection evaluations use this metric as it expresses the overlap between an object label $O$ and an object hypothesis $H$. A common choice of the minimum detection overlap is $\theta_{IOU}=0.5$~\cite{voc}. \\
 \textbf{Scaling Error:} Assuming case (a) with a perfectly positioned hypothesis but a scaling error $\mathcal{E}_k$, the intersection over union is given by  
 \begin{align}
 \theta_{\mathrm{IOU}}= \frac{w^\mathrm{p}_{\mathrm{O}}\cdot r\cdot w^\mathrm{p}_{\mathrm{O}}}{(1\pm\mathcal{E}_k)w^\mathrm{p}_{\mathrm{O}} \cdot r\cdot (1\pm\mathcal{E}_k)w^\mathrm{p}_{\mathrm{O}}} = \frac{1}{(1\pm\mathcal{E}_k)^2},
 \end{align}
 leading to
 \begin{align}
 |\mathcal{E}_k| =\frac{1}{\sqrt{\theta_{\mathrm{IOU}}}} -1.
\end{align}

\begin{figure}[!t]
\begin{center}
	\includegraphics[width=0.85\linewidth]{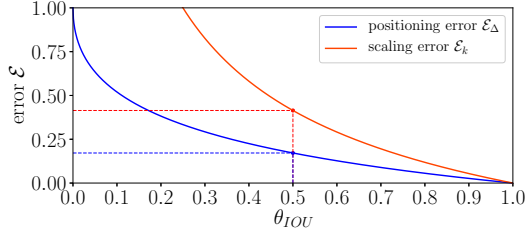}
\end{center}
	\caption[Scaling Factor]{Scaling error $\mathcal{E}_k$ and step size error $\mathcal{E}_\Delta$ for $0\leq \theta_{\mathrm{IOU}} \leq 1$. In order to reach an intersection over union of 0.5 the annotated object rectangle and the object hypothesis may differ by 41 percent in size. A step size error of 18 percent is allowed.}
	\label{fig:scaling_error}
\end{figure}

\textbf{Positioning Error:} Assuming case (b) with a perfectly sized hypothesis H, with a positioning error $\mathcal{E}_\Delta$ resulting from the window step size, the intersection over union can be formulated as

{\small
\begin{align}
\theta_{\mathrm{IOU}} = \frac{w^\mathrm{p}_{\mathrm{O}}(1-\mathcal{E}_\Delta)\cdot r\cdot w^\mathrm{p}_{\mathrm{O}}(1-\mathcal{E}_\Delta)}{w^\mathrm{p}_{\mathrm{O}}(1+\mathcal{E}_\Delta) \cdot r\cdot w^\mathrm{p}_{\mathrm{O}}(1+\mathcal{E}_\Delta)} = \frac{(1-\mathcal{E}_\Delta)^2}{(1+\mathcal{E}_\Delta)^2}.
\end{align}} %
The positioning error $\mathcal{E}_\Delta$ contains the information on the step size, so reformulating this equation yields a solvable quadratic equation. The solution is given by

\begin{align}
\label{eq:delta}
\mathcal{E}_\Delta  = \frac{\theta_{\mathrm{IOU}}- 2 \sqrt{\theta_{\mathrm{IOU}}}+1}{1-\theta_{\mathrm{IOU}}},
\end{align}

while the second solution is rejected because of inappropriate values.
Figure~\ref{fig:scaling_error} illustrates the relationship between the scaling error $\mathcal{E}_k$, positioning error $\mathcal{E}_\Delta$ and the intersection over union. As an operating point $\theta_{IOU}=0.5$ is also illustrated. Depending on the desired accuracy $\theta_{IOU}$ a scaling factor $k$ results which can be used to define the width-sum in Equation (\ref{eq:hyps}). In order to reach $\theta_{IOU}=0.5$ a step size error $\mathcal{E}_\Delta=0.18$ or scaling error $\mathcal{E}_k=0.41$ is allowed. All derivations are now combined and the resulting number of object candidates for each frame interpreted.

 \begin{figure}[!t]
 	\begin{center}
 	\includegraphics[width = 0.85\linewidth]{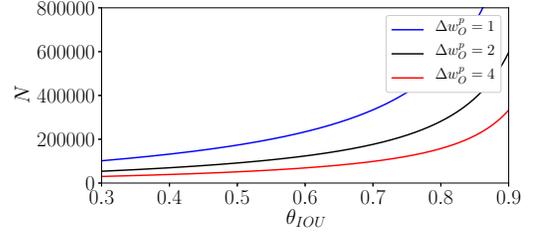}
 	\end{center}
 	\caption[Simulated number of Hypotheses per Frame]{Theoretical number of hypotheses $N$ with respect to the intersection over union $\theta_{\mathrm{IOU}}$.  Following assumptions are made: aspect ratio $r=3$, image size: 1242x375, $w^p_{O, min}=10$, $w^p_{O, max}=100$. High width step sizes decrease the number of hypotheses. At least 100 000 hypotheses are necessary to guarantee an overlap of 50 percent.  }
 	\label{fig:hyps_nr}
 \end{figure}

%\begin{figure}[H]
%\begin{center}
%	\includegraphics[width=1\linewidth]{fig/step_size_delta_plus.eps}
%\end{center}
%	\caption[Step Size Error - Solution 1]{Step size error for $0\leq \theta_{\mathrm{IOU}} \leq 1$. The relation between $\Delta^+$ and the intersection over union is approximately exponential. The step size error is normalized between 0 and 1. In order to guarantee a detection accuracy of $\theta_{IOU}=0.5$ a step size error of 10 percent is allowed which results in a step size of 20 percent ($2\cdot\Delta$)}
%	\label{fig:delta_plus}
%\end{figure}
% 

 \subsection{Interpretation}
Combining Equation (\ref{eq:hyps}), which defines the general number of hypotheses and (\ref{eq:delta}) as the allowed step size error leads to a final equation which can be simplified under the assumptions $w^\mathrm{p}_\mathrm{I} \gg w^\mathrm{p}_{\mathrm{O,}i}$ and $h^\mathrm{p}_\mathrm{I} \gg r\cdot w^\mathrm{p}_{\mathrm{O,}i}$ to
 
 {\footnotesize
 \begin{align}
 N\lesssim \mathlarger{\sum}\limits_{w^\mathrm{p}_{\mathrm{O,min}}}^{w^\mathrm{p}_{\mathrm{O,max}}} ~\mathlarger{\sum}\limits_{r_{\mathrm{min}}}^{r_{\mathrm{max}}} \frac{2\cdot (1-\theta_{\mathrm{IOU}})}{(\theta_{\mathrm{IOU}}+1)- 2 	\sqrt{\theta_{\mathrm{IOU}}}}\cdot \Bigg[\frac{w^\mathrm{p}_\mathrm{I} \cdot h^\mathrm{p}_\mathrm{I}}{r_i\cdot (w^\mathrm{p}_{\mathrm{O,}i})^2}\Bigg].
 \end{align}}%
 
 Figure~\ref{fig:hyps_nr} illustrates this equation for an assumed image size of $w^\mathrm{p}_\mathrm{I}=1242$ and $h^\mathrm{p}_\mathrm{I}=375$ pixels (KITTI resolution). Three different curves with different choices for the width step size are illustrated in the range $0.3 \leq \theta_{IOU} \leq 0.9$. For high intersection over union requirements, the number of hypotheses $N$ increases greatly. Furthermore, small objects to be detected $w^\mathrm{p}_{\mathrm{O,min}}$ also greatly increase $N$ as small objects are only detected by small step sizes.
 This theoretical investigation clearly shows that the conventional sliding window approach creates a huge number of object candidates per frame. Strict requirements in geometric accuracy $\theta_{IOU}$ and small objects to be detected increase $N$. A classification of millions of objects in real-time remains a challenging or impossible task despite current computing power. Using powerful classifiers and features is especially critical because of their computational cost. Motivated by this investigation, a sliding window approach is now presented, which takes the disparity image into account. Using depth information results in a major decrease in the number of object candidates and allows the usage of Equation (\ref{eq:delta}).

	  \begin{figure*}[!t]
		 	 	  	\begin{center}
		 	 	  		\includegraphics[width = 0.75\linewidth]{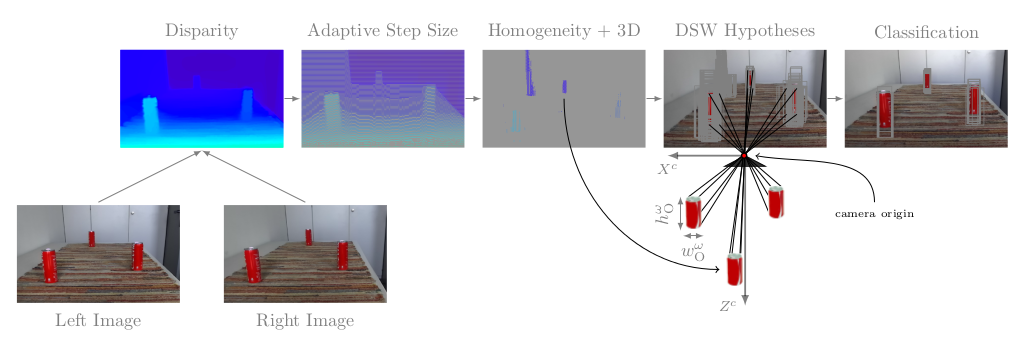}
		 	 	  	\end{center}
		 	 	  	\caption{Illustration of all steps of the proposed DSW approach for the detection of a beverage can. The disparity image is calculated via SGM algorithm. After applying the adaptive step size a sampled disparity image results, where close objects are sampled with a lower resolution. The choice of the step size guarantees a defined detection accuracy ($\theta_{IOU}$), independent of the objects distance. Especially the homogeneity verification is very effective in rejecting inappropriate object candidates regions. The disparity values of the desk and left wall decrease with depth, which results in a rejection of these disparity regions. Transition regions between beverage can and desk/wall are rejected as well, as disparity value jumps occur. The disparity hypotheses are then mapped into the color image. Applying the depth information results in perfectly sized hypotheses and avoids computationally expensive window scalings.    }
		 	 	  	\label{fig:principle}
		 	 	  \end{figure*}  
\section{Disparity Sliding Window (DSW)}
\label{sec:dsw}
The disparity sliding window method requires a disparity image as an input. Furthermore the stereo camera must be calibrated, i.e. intrinsic camera properties must be available. 
\subsection{Basic Principle}
\label{sec:bp}
Many object detection problems deal with the detection of objects of a fixed (or approximately) real-world size. Knowing the real-world size of the object as well as its distance to the camera origin clearly describes it in three-dimensional space. Combining this knowledge with the intrinsic parameters, the projection of an object on the image plane can be calculated. This idea is used by the proposed DSW approach.
Instead of creating many computational expensive window scalings at a given position (pixel) in the image, a single hypothesis is created via projection. The object is defined by its real-world width $w^\omega_O$ and height $h^\omega_O$. $\omega$ denotes world coordinates. The distance component $Z^c$ in the camera coordinate frame is received from the disparity image. This leads to the definition of four real-world corner points for the object 

{\footnotesize
\begin{align}
\hat{\underline{X}}^\mathrm{c}_{\mathrm{ll}} =  \begin{bmatrix} 0 \\ 0 \\Z^\mathrm{c}  \end{bmatrix}, 
\hat{\underline{X}}^\mathrm{c}_{\mathrm{lr}} =   \begin{bmatrix} w^\omega_O\\ 0 \\Z^\mathrm{c},  \end{bmatrix},  \hat{\underline{X}}^\mathrm{c}_{\mathrm{ur}} =   \begin{bmatrix} w^\omega_O \\ h^\omega_O \\Z^\mathrm{c}  \end{bmatrix}, 
\hat{\underline{X}}^\mathrm{c}_{\mathrm{ul}} =  \begin{bmatrix} 0 \\ h^\omega_O \\Z^\mathrm{c} \end{bmatrix}.  
\end{align}} %

given as lower left, lower right, upper left and upper right corner. These points circumscribe the object from the camera point of view.
Reprojecting those 3D points 

\begin{align}
\hat{\underline{X}}^\mathrm{c}_{\mathrm{ll}} \rightarrow \hat{\underline{x}}^\mathrm{p}_{\mathrm{ll}} \qquad \hat{\underline{X}}^\mathrm{c}_{\mathrm{ur}} \rightarrow \hat{\underline{x}}^\mathrm{p}_{\mathrm{ur}},
\end{align} 

gives the corresponding corner points on the image plane. The pinhole camera model~\cite{Hartley2004} is used, which describes the mapping of a 3D point on a 2D image plane as

\begin{align}
\underline{\hat{x}}^\mathrm{p}=\mathbf{K}~[\mathbf{I}_{3x3}|\underline{0}]~\underline{\hat{X}}^\mathrm{c}.
\end{align} 
$\mathbf{K}$ is the intrinsic camera matrix defined by focal lengths, skew parameter and principal point.
 A vector subtraction of lower left and upper right corner gives the resulting width $ w^\mathrm{p}_{\mathrm{O}}$ and height $ h^\mathrm{p}_{\mathrm{O}}$ of the hypothesis at a given disparity pixel

\begin{align}
 \begin{bmatrix} w^\mathrm{p}_{\mathrm{O}} & h^\mathrm{p}_{\mathrm{O}}   \end{bmatrix}^T = \hat{\underline{x}}^\mathrm{p}_{\mathrm{ur}} -\hat{\underline{x}}^\mathrm{p}_{\mathrm{ll}}.
\end{align}

 This projection is calculated at each pixel in the disparity image, resulting in one window or hypothesis size per pixel. A conventional sliding window approach would create a high number of windows per pixel because the size of the object at the current position is unknown.\\
\textbf{Speed-Up:} A projection of a high number of points is computationally expensive due to matrix multiplication for each point (intrinsic camera matrix). Since a disparity image has a defined resolution with values $d_{min}$ up to $d_{max}$ and quantization $\delta_d$, the object width and height on image plane can be precomputed for each possible disparity value. A projection is not needed anymore and is replaced by a fast table access. Table~\ref{tab:lut} illustrates the LUT, which can be used for fast calculation of the hypotheses width and height with respect to the disparity value.
	\begin{table}[H]
		\centering
		\caption{Lookup table for fast determination of the object width and height for a given disparity.  } \label{tab:lut}
		\scalebox{0.8}{
		\begin{tabular}
			{lcc} \toprule  \textbf{disparity d}& $w^\mathrm{p}_{\mathrm{O}}$ & $h^\mathrm{p}_{\mathrm{O}}$ \\
			\midrule
			$d_{\mathrm{min}}$ & ... & ...  \\
			$d_{\mathrm{min}}$ + $\delta_d$ & ... & ...  \\
			$d_{\mathrm{min}}$ + $2\delta_d$ & ... & ...  \\
			\vdots & \vdots & \vdots  \\
			$d_{\mathrm{max}}$& ... & ...  \\
			\bottomrule
		\end{tabular}}
	\end{table}

	\subsection{Adaptive Step Size}
	
	Conventional sliding-window approaches typically use fixed predefined window step sizes, which form a good compromise between the detection accuracy and the number of candidates per frame. Especially for distant objects, high step sizes can be critical. We now propose an adaptive step size which is dependent on the disparity values of an object. As a consequence, a small step size for distant objects and a high step size for close objects is used. The calculated object width $w^\mathrm{p}_{\mathrm{O}}$ in pixels is used to calculate the step sizes

	\begin{align}
	s^\mathrm{p}_\mathrm{x} = \Delta \cdot w^\mathrm{p}_\mathrm{O}=\frac{2(\theta_{\mathrm{IOU}}+1)- 4 \sqrt{\theta_{\mathrm{IOU}}}}{1-\theta_{\mathrm{IOU}}}\cdot w^\mathrm{p}_\mathrm{O}
	\label{eq:ss_x}
	\end{align} 
	and 
	\begin{align}
	s^\mathrm{p}_\mathrm{y} = \Delta \cdot  h^\mathrm{p}_\mathrm{O}= \frac{2(\theta_{\mathrm{IOU}}+1)- 4 \sqrt{\theta_{\mathrm{IOU}}}}{1-\theta_{\mathrm{IOU}}}\cdot r\cdot w^\mathrm{p}_\mathrm{O}
	\label{eq:ss_y}
	\end{align} 
	with $s^\mathrm{p}_\mathrm{x}$ and $s^\mathrm{p}_\mathrm{y}$ given as step size in $x^\mathrm{p}$ and $y^\mathrm{p}$ direction, respectively. Again an aspect ratio of $r$ is assumed. This adaptive step size guarantees a defined detection accuracy $\theta_{IOU}$ determined in Section~\ref{sec:csw}. The choice of $\Delta=2\mathcal{E}_\Delta$ is inspired by Equation (\ref{eq:delta}) and Figure~\ref{fig:scaling_error}. Please note, that for practical use of this equation decimal numbers are rounded. Disparity jumps occur at transition areas of objects with different distance to the camera origin. Especially for transitions from close to far objects, high step sizes can cause a jump over an existing small object. We therefore propose to verify disparity jumps and appropriately correct the step size, so that a jump over existing objects is avoided.     
	
	\subsection{Impact of Disparity Inaccuracy}
	
	A drawback of calculating depth via stereo is the rising inaccuracy with distance. This inaccuracy does not adversely affect the DSW method. The width of an object in pixels can be formulated as a subtraction of two 3D points 
	
	\begin{align}
	w^\mathrm{p}_{\mathrm{O}}=x^\mathrm{p}_1 - x^\mathrm{p}_2 =\frac{f}{Z^\mathrm{c}} (X^\mathrm{c}_1 - X^\mathrm{c}_2).
	\end{align} 
	
\begin{figure}[!t]
		 	\begin{center}
		 		\includegraphics[width=0.65\linewidth]{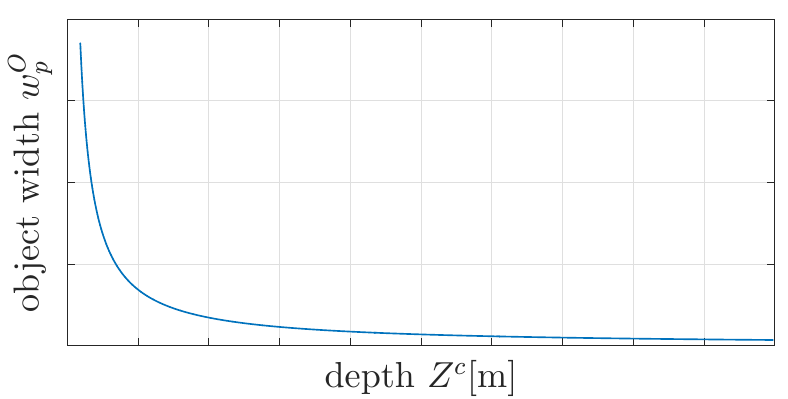}
		 	\end{center}
		 	\caption{Relationship between object width in pixels with respect to the real-world distance of the object. The behavior counteracts the rising depth inaccuracy of stereo algorithms.}
		 	\label{fig:stereo_inaccuracy}
		 \end{figure}

	 It follows, that the width in pixels is inversely proportional to the distance of the object to the camera origin ($w^\mathrm{p}_{\mathrm{O}}  \sim \frac{1}{Z^\mathrm{c}}$). 
	 Figure~\ref{fig:stereo_inaccuracy} illustrates the relationship between object width in pixel and the distance of the object in meters. It can be seen, that the gradient strongly decreases with increasing distance 
	 
	 \begin{align}
	 \frac{\partial w_\mathrm{O}^p}{\partial Z^c}\sim -\frac{1}{(Z^c)^2}.
	 \end{align}
	 
	 Since the inaccuracy of depth measurement received by disparity images increases with depth both gradients compensate. A depth error of 20m measured for an object at $Z^c\gg100\mathrm{m}$ hardly changes the object width in pixels. However, quantization problems are more critical for distant objects. Since only integer pixels are processed, rounding errors increase with the objects distance.

	 \subsection{Homogeneity Verification}
	 
	A further method for reduction of the number of candidates with additional disparity information is a \emph{homogeneity verification} through which the condition that most objects to be detected are upright, i.e. have an approximately constant disparity distribution in the image, is verified. One possibility to verify the homogeneity is a standard deviation calculation of all pixels $\underline{x}^\mathrm{p}$ inside the object candidate $\sqrt{\mathrm{Var}(\underline{x}^p)}$. 	 	
		Only object hypotheses with a standard deviation smaller than $\sigma$ are further processed. Since a standard deviation calculation based on all pixels of a high number of objects is computationally expensive, only several points are used for calculation. This merged as a good compromise between performance and computational cost. Depending on the shape of the object, the points used for the calculation have to be distributed. Edge regions should be avoided as stereo algorithms tend to be error-prone at object transitions.

	 \subsection{3D Region of Interest}
	 In most object detection cases objects only appear in a certain three dimensional region. Furthermore for many applications it is only necessary to detect objects up to a defined distance. Conventional sliding window approaches also create and classify object hypotheses in regions which are irrelevant for the given task.
	 Since the DSW approach uses disparity information each pixel of the color image can be projected into 3D space. A set $S$ of 3D world points $S = \{\underline{\hat{X}}^c_1, \underline{\hat{X}}^c_2, ..., \underline{\hat{X}}^c_n\}$
	 results after appending the pinhole camera model. After applying 3D constraints a filtered set of points $S_f$ results, which can be formulated as 
	 
	 $	 S_{\mathrm{f}} = \Bigg\{\underline{\hat{X}}^c_n \Bigg| \underline{X}^c_{min}  \leq \underline{\hat{X}}^c_n \leq  \underline{X}^c_{max}\Bigg\},
	 $
	   where $\underline{X}^c_{min} = \begin{bmatrix} X^c_{\mathrm{min}} & Y^c_{\mathrm{min}} & Z^c_{\mathrm{min}} \end{bmatrix}^T$ and $\underline{X}^c_{max} = \begin{bmatrix} X^c_{\mathrm{max}} & Y^c_{\mathrm{max}} & Z^c_{\mathrm{max}} \end{bmatrix}^T$ .
	   Remapping the remaining 3D points on the image gives a filtered disparity image. Only remaining pixels are further processed by DSW resulting in a further reduction of the number of candidates.

%	 \subsection{Remapping Disparity Hypotheses on the Unrectified Color Image}
%	 
%	 All steps of DSW previously described are applied on the disparity image. Following classification steps are generally used in combination with the raw color image. Consequently, each disparity hypothesis has to be unrectified. 
%	 This step can be omitted when classifying with a rectified color image.\\
%	 Figure~\ref{fig:principle} illustrates all steps of the DSW approach by the example of detecting a coke can.

\section{Experiments}
\label{sec:application}
In this section we evaluate the DSW approach on the KITTI~\cite{Geiger2012} dataset. We only evaluate our method as an object proposal algorithm without a subsequent classification. KITTI provides accurate 3D and 2D annotations of different classes, such as cars, vans, cyclists or pedestrians. Since our method is designed to predict 2D image detections, we only evaluate our method using 2D annotations.
We want to clarify the following properties of our approach:\\
\textbf{Quality of Proposals:} The quality of proposals is expressed by accuracy in position and size. We again use the metric intersection over union (see Equation~(\ref{iou_theta})).\\
\textbf{Quantity of Proposals:} In order to express the quantity of proposals we measure the number of proposals per image (PPI). This number is crucial for the detection algorithm, as it mainly defines the overall execution time.\\
\textbf{Parameterization Impact:} As each algorithm, the DSW approach has several parameters to choose. Model size, adaptive step sizes (determined by the desired minimum IoU), 3D regions and the homogeneity threshold have to be defined. Our results clarify the impact of those parameters on both, quality and quantity of proposals. 
\begin{table} [!b]

\caption{Statistics of the pedestrians in KITTI used for evaluation}
\centering
\begin{tabular}{lllll} 
\toprule
&&\multicolumn{3}{c}{\textbf{occluded}}\\  \cmidrule{3-5}
\textbf{visibility} & \textbf{fully visible} & partly & largely & unknown\\  
\midrule 
\textbf{\# pedestrians} & 2667 & 1095 & 671 & 54\\ 

\bottomrule
\end{tabular}
\label{tab:kitti_ped}
\end{table}

  \begin{figure}[!b]
		 	 	  	\begin{center}
		 	 	  		\includegraphics[width = 0.75\linewidth]{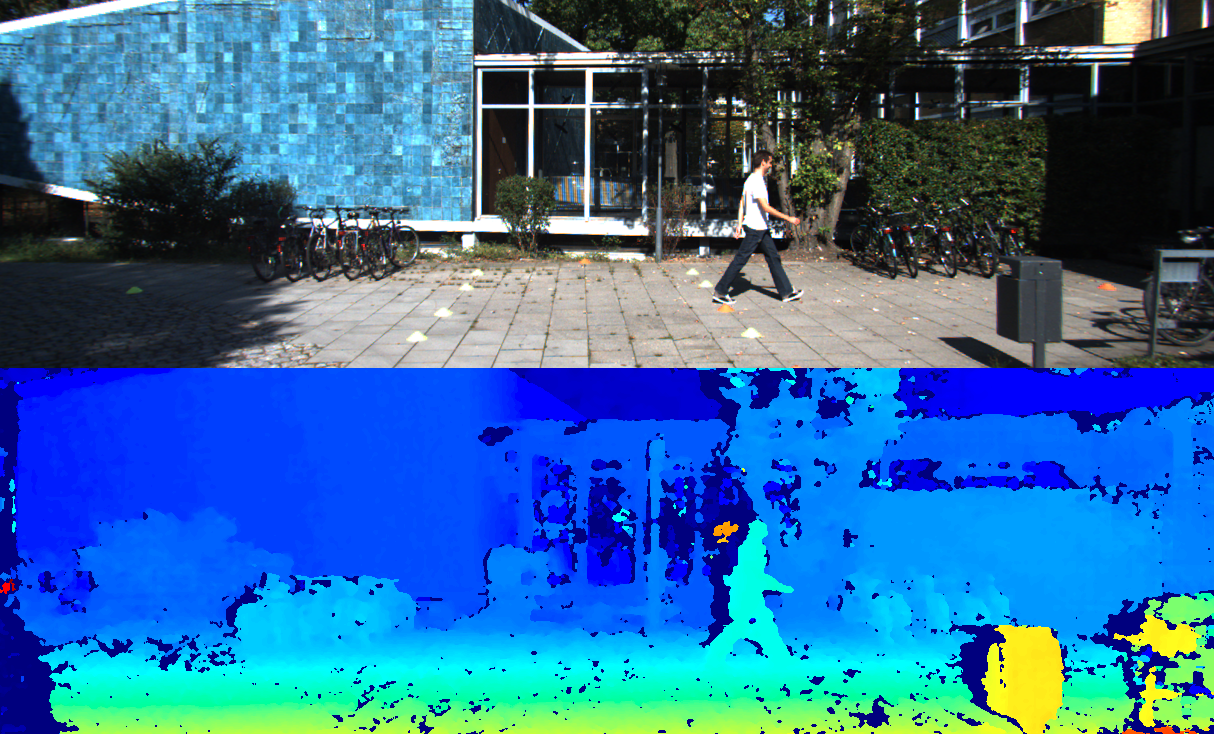}
		 	 	  	\end{center}
		 	 	  	\caption{Example of a disparity image calculated from left and right image of the KITTI object detection dataset.}
		 	 	  	\label{fig:kitti_disparity}
\end{figure}
 
\subsection{Dataset}
The KITTI object detection benchmark provides 2D bounding box annotations. Furthermore, left and right camera image (rectified) as well as calibration data is provided. We evaluate our method on the class pedestrian. Table~\ref{tab:kitti_ped} shows the statistics of the training dataset used for evaluation. The dataset contains 4487 annotated pedestrians, from which 1766 are partly or largely occluded.

 \begin{figure*}[!t]
         \begin{subfigure}[b]{0.32\textwidth}
                 \includegraphics[width=\linewidth]{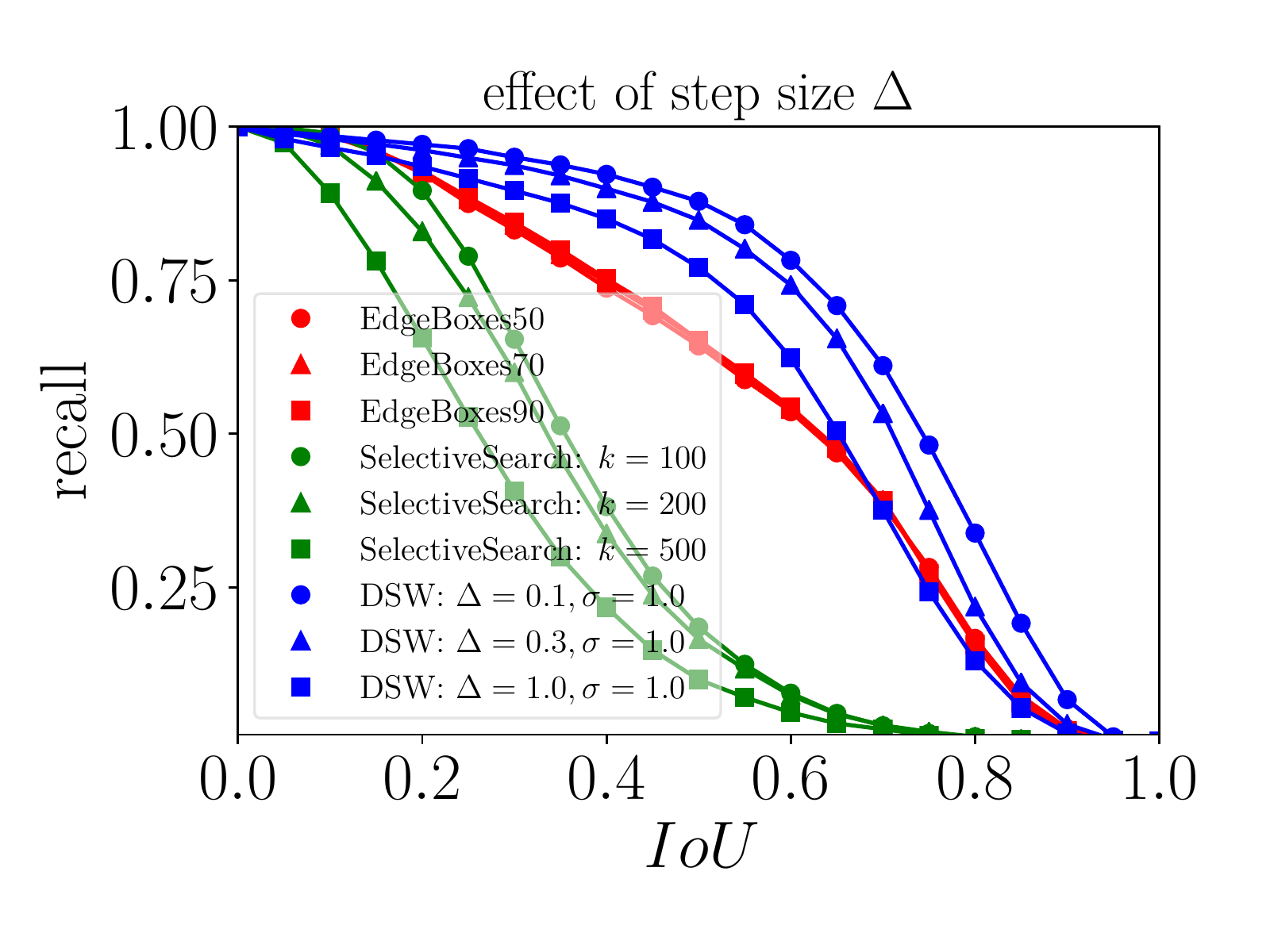}
                 \caption{Quality w.r.t. $\Delta$}
                 \label{fig:gull}
         \end{subfigure}%
         \begin{subfigure}[b]{0.32\textwidth}
                 \includegraphics[width=\linewidth]{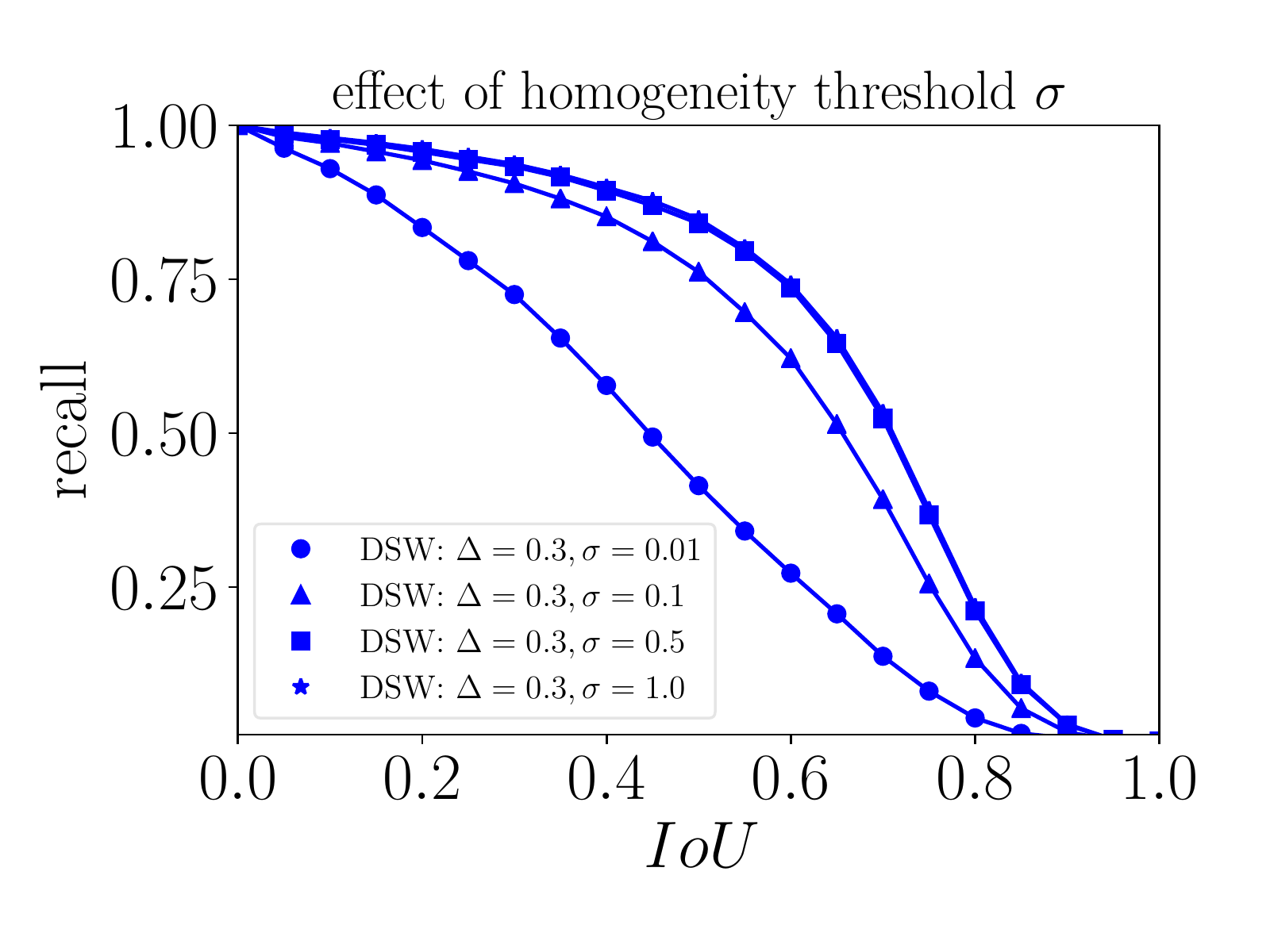}
                 \caption{Quality w.r.t. $\sigma$}
                 \label{fig:gull2}
         \end{subfigure}%
         \begin{subfigure}[b]{0.32\textwidth}
                 \includegraphics[width=\linewidth]{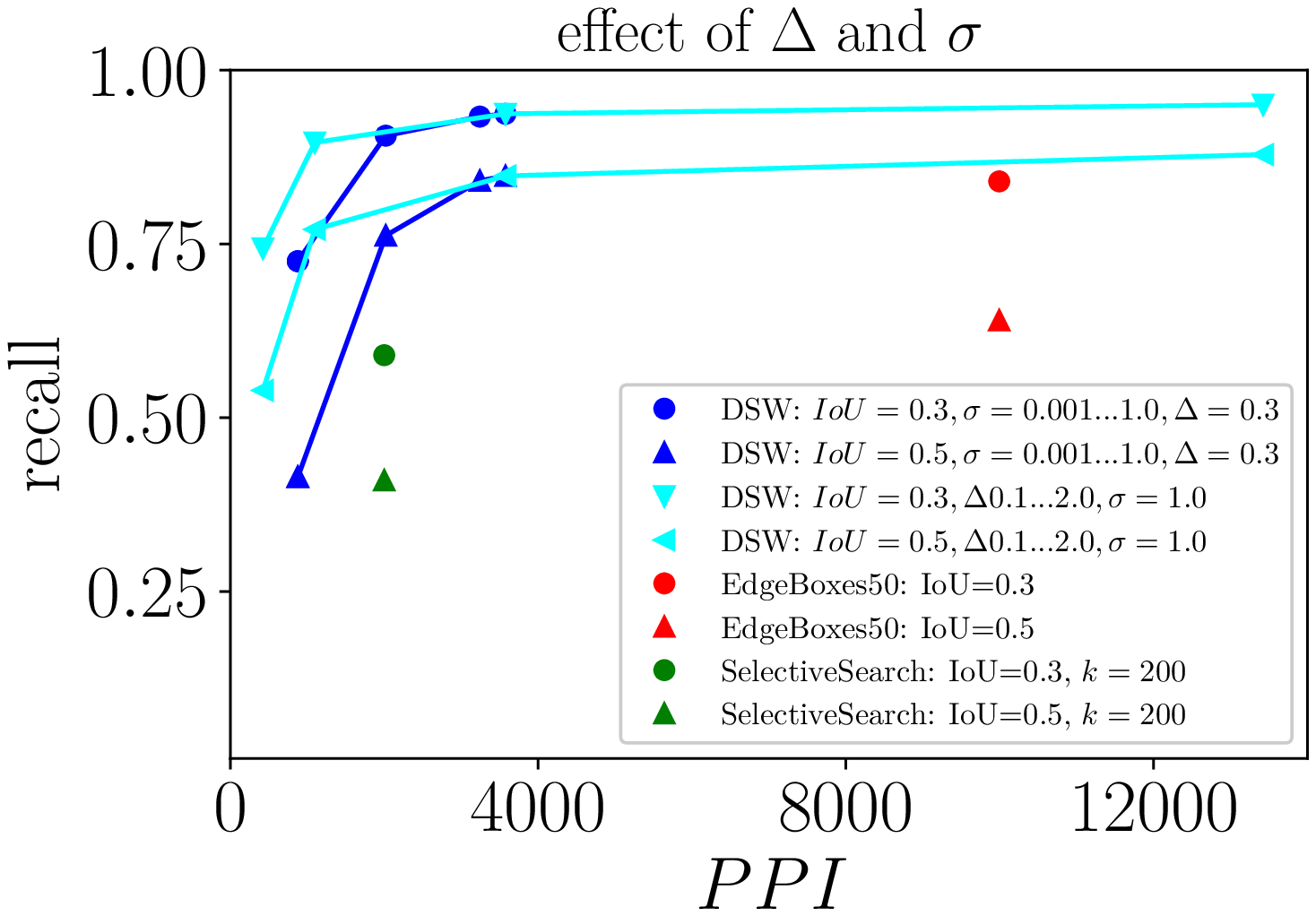}
                 \caption{Quantity w.r.t. $\Delta$ and $\sigma$}
                 \label{fig:tiger}
         \end{subfigure}%
         \caption{Evaluation results of DSW for all pedestrians (including occluded pedestrians). The left and middle figure illustrate the recall of our algorithm with respect to the overlap to the ground truth. Different curves for several choices of adaptive step size and minimum homogeneity threshold value are shown. The right figure illustrates the recall with respect to the average number of proposals per image. Lower step sizes and higher homogeneity threshold values increase the quality but also the number of proposals.}\label{fig:eval}
 \end{figure*}

\textbf{Disparity calculation:} KITTI does not provide disparity images. We use the Semi-Global Matching (SGM)~\cite{H.Hirschmuller2005} approach for disparity calculation. SGM is a good compromise between speed and accuracy. It is often used in intelligent vehicles context \cite{Gehrig2012} due to its real-time capability. Figure~\ref{fig:kitti_disparity} illustrates an example of a calculated disparity image.
\subsection{Choice of Parameters}
\textbf{Model Size}: Our approach is dependent on the choice of a real-world model size as introduced in Section~\ref{sec:dsw}. In order to detect objects, which vary in size, either multiple models have to be chosen or an average model has to be defined. The latter variant can lead to inaccurate bounding boxes depending on the deviation of the object to the average model. The first variant leads to a much higher number of proposals per image. We decided for an average pedestrian size of $w^{\omega}_O=0.60\mskip\thinmuskip \mathrm{m}$ and $h^{\omega}_O=1.73\mskip\thinmuskip \mathrm{m}$.\\ 
\textbf{Adaptive Step Size}: This parameter is crucial for the quality of the approach in terms of the IoU. According to Equations~(\ref{eq:ss_x}) and (\ref{eq:ss_y}) and Figure~\ref{fig:scaling_error}, $2\cdot\mathcal{E}_\Delta(0.5) \approxeq 0.3-0.4$ is a good step size choice. When assuming an additional scaling error (imperfect model, imperfect disparity), a step size of 30 percent is an appropriate value to detect objects with at least 50 percent overlap. Nevertheless, we evaluate the DSW approach for different step sizes.\\
\textbf{Homogeneity Threshold:} This parameter mainly does not influence proposal quality (IoU) but the quantity and recall of the approach. A good choice is 0.1 for pedestrians, but it may vary for other applications.\\
\textbf{3D Constraints:} To allow a fairer comparison with Selective Search we avoid 3D filtering in this evaluation. However, it is very effective for reducing the proposal quantity (e.g. no assumed pedestrian below ground or above a certain height).
\subsection{Quality of Proposals}

\begin{figure*}[!t]
		 	 	  	\begin{center}
		 	 	  		\includegraphics[width = 0.95\linewidth]{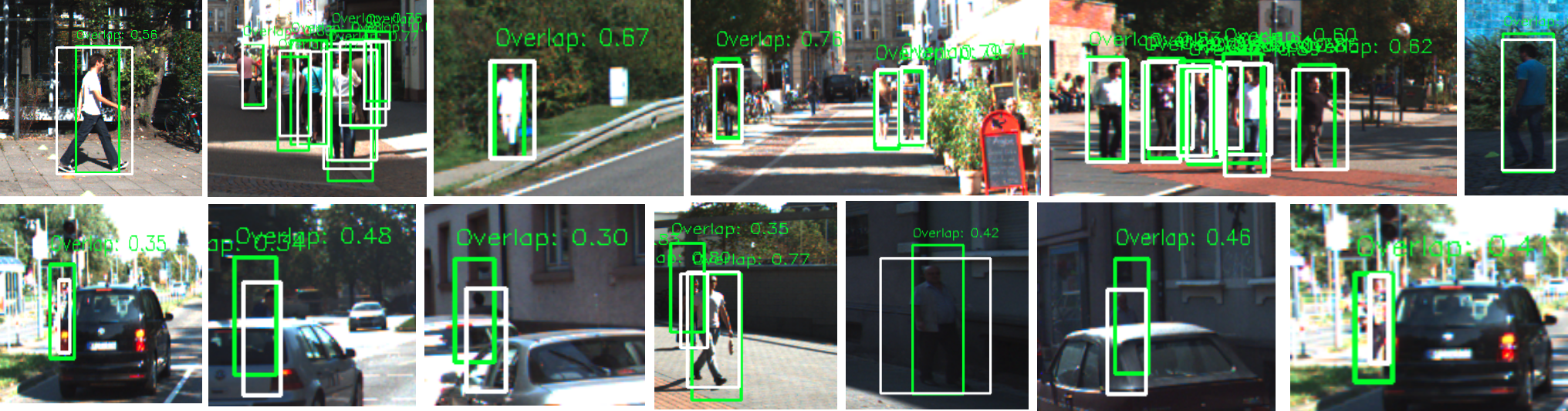}
		 	 	  	\end{center}
		 	 	  	\caption{Detection results (best overlap only) for class pedestrian on the KITTI object detection benchmark (green detection, white label). The first row illustrates detection results for $IoU>0.5$. Even groups of pedestrians are successfully detected. The model assumption fits well, as the to be detected objects are circumscribed by appropriate bounding boxes. The lower row illustrates cases with poor overlaps. They are mainly caused by occlusion, which lead to errors in position and size due to wrong disparity values. Annotation inaccuracy (see example 5, row 2) also causes low IoU values.  }
		 	 	  	\label{fig:det_examples}
\end{figure*}

As already defined, we measure the quality of proposals by its IoU value. Figures~\ref{fig:eval} (a) and (b) illustrate the recall of our DSW algorithm with respect to the IoU. Recall is defined as 
\begin{align}
\text{recall} = \frac{\mathrm{TP}}{\mathrm{TP}+\mathrm{FN}},
\end{align}
 whereas a true positive (TP) has to have a larger overlap than IoU with one annotation. False negatives (FN) are annotations not covered by at least one hypotheses with the given overlap. Different operating points of the algorithm (increasing adaptive step size and homogeneity threshold) are evaluated. We reach a recall of approximately 85 percent with an overlap criteria of 0.5 and 4000 proposals per image. We reach recall values close to 80 percent when only creating around 1000 proposals per image. Example detections can be seen in Figure~\ref{fig:det_examples} as well as poor overlaps due to occlusion or annotation inaccuracies. High recall for high overlap criteria can be reached by decreasing the adaptive step size. 

 \subsection{Quantity of Proposals}
 
 Figure~\ref{fig:eval}(c) illustrates the quantity of proposals expressed as a recall ROC curve with respect to the average number of false positives per frame. The strength of sliding-window approaches are typically high recall values at the expense of clearly higher false positive rates. Good operating points are between 500 and 3000 proposals per image. 
\subsection{Comparison With State-of the Art}
We evaluate our method against the state-of-the-art methods \emph{Selective Search}~\cite{Uijlings2012} and EdgeBoxes~\cite{Zitnick2014}.\\
\textbf{Selective Search:} Figure~\ref{fig:eval}(a) illustrates the results for Selective Search (green lines) for different parameter setting on the KITTI object detection benchmark for the class pedestrians. DSW outperforms Selective Search by 26 percent for $\theta_{IOU}=0.3$ and 70 percent for $\theta_{IOU}=0.5$ . Selective Search detects objects from similarity in color, texture, size and gap filling. The appearance of pedestrians is highly diverse. Furthermore, the objects in KITTI are rather small compared to other benchmarks, such as PASCAL VOC.  Therefore, the approach cannot perform well. Nevertheless, Selective Search is a class-agnostic approach distinguished for high recall on other object detection problems with only creating a small number of proposals per image.       \\
\textbf{EdgeBoxes:} Figure~\ref{fig:eval}(a) illustrates the results for EdgeBoxes (red lines). We follow the parameterization of the original paper and evaluate EdgeBoxes50 to EdgeBoxes90 with $\delta=0.5, 0.7, 0.9$. EdgeBoxes reaches a higher recall than selective search, mainly caused by its independence on appearance. Nevertheless, DSW also outperforms EdgeBoxes by 9 percent for $\theta_{IOU}=0.3$ and 19 percent for $\theta_{IOU}=0.5$. Note, that we allowed 10\,000 proposals per image for EdgeBoxes. 

\begin{table}[!t]
\centering
\caption{Recall of all three approaches for different parameters and IoU thresholds.}
\scalebox{0.75}{
\begin{tabular}{l|ccc|cc|ccc} 
\toprule
&  \multicolumn{3}{c}{\textbf{Ours (DSW)}}  &   \multicolumn{2}{c}{\textbf{Selective Search}} &  \multicolumn{3}{c}{\textbf{EdgeBoxes}}\\  

&$\Delta$: 0.1 & 0.3 & 0.5 & $k$: 200 & 500 & $\delta$: 0.5 & 0.7 & 0.9\\ 
\midrule 
\textbf{recall} ($\theta_{IOU}=0.3$)& \textbf{0.95}  & \textbf{0.94} & \textbf{0.89}& 0.59 & 0.41 & 0.84 & 0.85 &0.85 \\ 
\midrule
\textbf{recall} ($\theta_{IOU}=0.5$) & \textbf{0.88} & \textbf{0.85} & \textbf{0.77} & 0.17 & 0.10& 0.64 & 0.66 & 0.66 \\
\bottomrule
\end{tabular}}
\label{tab:runtime}
\end{table}
\textbf{Runtime Comparison:} Table~\ref{tab:runtime} shows the execution time of DSW for different choices of the adaptive step size. DSW requires 3-10 milliseconds per image, which makes the approach real-time capable. It is faster than Selective Search by a factor of 974 and faster than EdgeBoxes by a factor of 1537. Disparity calculation is not included in the runtime calculation. Each runtime is the average of all images.
\begin{table}[H]
\centering
\caption{Runtime comparison between DSW, Selective Search and EdgeBoxes.}
\scalebox{0.75}{
\begin{tabular}{llll} 
\toprule
&  \textbf{Ours (DSW)} &  \textbf{Selective Search} & \textbf{EdgeBoxes}\\  
\midrule 
\textbf{params} & $\Delta=[0.1, 0.3, 0.5]$ & $k=[200, 500]$ & $\delta=[0.5, 0.7, 0.9]$\\ 
\midrule
\textbf{runtime[s]} & [0.010, 0.0039, 0.0027] & [4.22, 2.63] & [4.95, 4.40, 4.15]\\
\bottomrule
\end{tabular}}
\label{tab:runtime}
\end{table}

\subsection{Limitations and Applications of DSW}
As already indicated DSW is particularly predestined for single-class problems, where prior knowledge of the object is given. Challenges such as the PASCAL VOC challenge, which need to detect a high number of different object classes, are not well suited for DSW. However, there are still many use cases for object detection of an single object such as autonomous driving (pedestrian, cyclist, traffic sign, traffic light detection) or robotics (detection of known and static objects in industrial or domestic environment). These use cases also often require real-time detection, which is enabled by DSW. Furthermore, these approaches work on stereo data anyhow. 
\section{Conclusion}
This paper proposes an algorithm for creating object proposals from disparity images in a sliding-window fashion. It overcomes the drawbacks of conventional sliding-window technique by reducing the overall number of proposals per image. An intelligent, adaptive step size helps to accurately detect small and large objects with help of the disparity image. Furthermore, DSW avoids multiple proposal sizes at each sliding-window position by calculating the bounding box size from projection with an assumed model size.
This characteristic makes the approach predestined for objects with a fixed or approximately fixed real-world size. Nevertheless, using multiple models allows to also use this approach for objects with a non-fixed real-world size.\\
DSW outperforms Selective Search and EdgeBoxes in recall. Furthermore, DSW is two decades faster. The execution time of a few milliseconds only makes the approach real-time capable for most applications.

\bibliographystyle{ieee}
\bibliography{egbib}

\end{document}